\documentclass[conference]{IEEEtran}
\IEEEoverridecommandlockouts

\usepackage{cite}
\usepackage{amsmath,amssymb,amsfonts}
\usepackage{algorithmic}
\usepackage{graphicx}
\usepackage{textcomp}
\usepackage{xcolor}
\usepackage{booktabs}

\usepackage{graphicx}
\usepackage{subcaption}
\usepackage{comment}
\usepackage{multirow}
\usepackage{url}
\usepackage{fancyhdr}
\fancypagestyle{draftnotice}{%
  \fancyhf{}%
  \fancyhead[C]{\parbox{\textwidth}{\centering\footnotesize\itshape This is the authors' draft version of the paper. For the final version of record, please refer to the official version published in the IEEE Intelligent Vehicles Symposium (IV), 2026, DOI: 10.1109/IV66570.2026.11624050.}}
}
\def\BibTeX{{\rm B\kern-.05em{\sc i\kern-.025em b}\kern-.08em
    T\kern-.1667em\lower.7ex\hbox{E}\kern-.125emX}}

\newcommand{\etal}{et al.}

\begin{document}

\title{Difference Feature Map Distillation: Transferring Inter-Sample Relational Knowledge Towards Efficient Transformer-Based Tracking
\\
}

\author{Zhicheng Ding$^{1}$, Xinyu Chu$^{1}$, Qing Tian$^{2*}$
\thanks{*This work was supported by the National Science Foundation (NSF) under Award No. 2153404 and No. 2412285. (Corresponding author: Qing Tian.)}
\thanks{$^{1}$Ding and Chu are with the Department of Computer Science, Bowling Green State University, OH 43402, USA.
        {\tt\small \{dingz, xchu\}@bgsu.edu}}%
\thanks{$^{2}$Tian is with the Department of Computer Science, University of Alabama at Birmingham, AL 35294, USA.
        {\tt\small  qtian@uab.edu}}%
}


\maketitle
\thispagestyle{draftnotice}

\begin{abstract}
In autonomous driving perception, visual object tracking systems must satisfy stringent latency and power constraints while remaining robust in complex and dynamic environments. Although transformer-based trackers achieve state-of-the-art accuracy, their substantial computational and memory overheads hinder deployment on real-time, resource-constrained platforms. To move toward this goal, we propose Difference Feature Map Knowledge Distillation (DFM-KD), a novel relational distillation framework tailored for transformer-based visual object tracking. Unlike conventional feature distillation methods that minimize point-wise discrepancies (e.g., mean squared error) between teacher and student feature representations, DFM-KD transfers knowledge through inter-sample feature differences, explicitly aligning the relational structure of the feature space. By distilling how the teacher models appearance variation and consistency across samples, rather than enforcing similarity in absolute activations, DFM-KD enables the student to better capture the structural dynamics of visual changes within a batch. As a result, the distilled model exhibits enhanced feature robustness and improved tracking performance. Extensive experiments demonstrate that DFM-KD consistently outperforms conventional feature-level distillation methods in both tracking precision and success rates.
\end{abstract}


\section{Introduction}
Visual Object Tracking (VOT) is a fundamental perception task in autonomous driving, enabling systems to maintain continuous localization of surrounding agents such as pedestrians, cyclists, and vehicles. Reliable tracking supports a broad range of safety-critical functions, including interaction prediction, lane-change planning, collision avoidance, and follow-vehicle control. In these scenarios, onboard tracking modules must operate under strict computational, thermal, and latency budgets while maintaining high robustness in dynamically changing environments.
Recent advances in VOT have been driven by deep Transformer-based approaches \cite{cui2023mixformerv2} \cite{ye2022ostrack} \cite{cui2022mixformer}, which achieve state-of-the-art performance through expressive token interactions and powerful cross-attention mechanisms. However, these performance gains come at the cost of substantial computation and memory usage, limiting their applicability for real-time, resource-constrained platforms commonly found on autonomous vehicles. A practical solution is to compress a high-capacity Transformer into a lightweight student model via knowledge distillation (KD), allowing the student to inherit the teacher’s accuracy while reducing inference cost.

\begin{figure}[t]
\centerline{\includegraphics[width=0.8\linewidth]{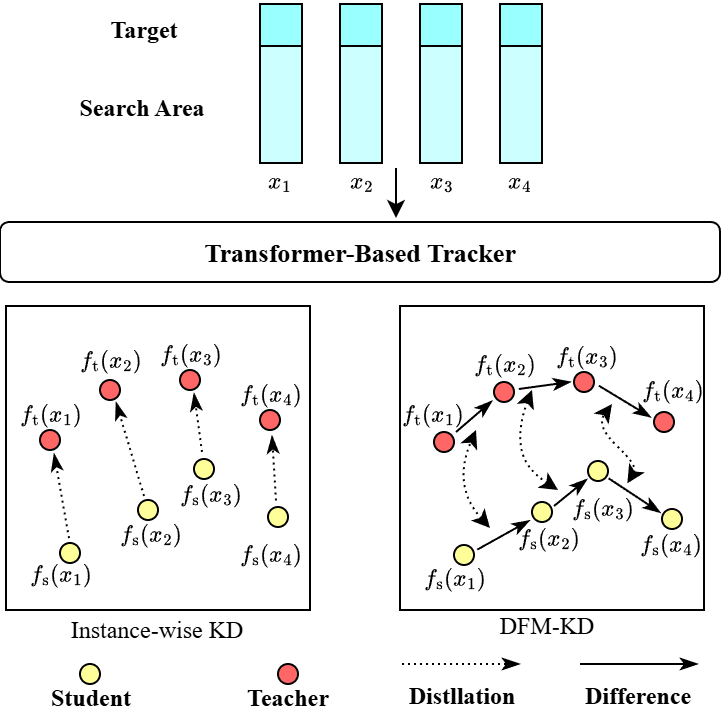}}
\caption{General idea of Difference Feature Map Distillation (DFM-KD). Unlike classic instance-wise KD, which aligns teacher and student features for each sample independently, DFM-KD distills the \emph{differences} between samples to preserve the teacher’s batch-wise topology. In the left panel, single-headed dotted arrows depict conventional per-sample alignment. In the right panel, DFM-KD introduces relational alignment: double-headed arrows represent the alignment of difference vectors between neighboring samples, enabling the student to capture how the teacher’s representations vary across the batch rather than matching isolated feature points.
}
\label{fig:dfmkd}
\end{figure}

Existing KD strategies in visual tracking, however, remain primarily instance-wise. A representative milestone in VOT distillation is MixFormerV2 \cite{cui2023mixformerv2}, which adopts a deep-to-shallow distillation paradigm. More recently, TDSC-KD \cite{11097763} introduces target-driven and student-centered distillation, reducing over-reliance on the teacher. Despite their differences, these approaches share the same underlying paradigm: they encourage the student to mimic the teacher’s absolute feature maps for each sample independently.
This instance-wise imitation paradigm overlooks a fundamental aspect of what makes a powerful teacher model effective. Beyond producing discriminative representations for individual samples, a well-trained teacher implicitly organizes samples into a meaningful batch-wise topology. For example, two template–search pairs depicting visually similar targets tend to be mapped closer in the teacher’s latent space, whereas pairs containing different targets are mapped farther apart. These relative arrangements reflect how the teacher encodes inter-sample relationships, an important component of its “dark knowledge” that is entirely discarded when performing point-wise distillation.

To address this limitation, we propose Difference Feature Map Distillation (DFM-KD), a new distillation paradigm designed to transfer inter-sample relational structure rather than relying solely on absolute feature reconstruction. As illustrated in Fig.~\ref{fig:dfmkd}, instead of enforcing $f_s(x_i) \approx f_t(x_i)$ for each sample individually, DFM-KD aligns the teacher’s and student’s difference feature maps, which capture how the teacher’s representations vary across different samples within a batch. These difference tensors encode appearance variation, contextual similarity, and structural relationships between samples — information unavailable in conventional feature-level distillation.
DFM-KD provides a stronger form of topological guidance: it encourages the student to preserve the teacher’s batch-wise manifold structure, thereby improving its ability to separate different targets from distractors and to maintain robustness under pose changes, illumination variations, and occlusions.
It enforces the student to match the relationships encoded by the teacher between arbitrary sample pairs $(x_i, x_j)$, aligning the overall structure of the student’s feature-space manifold with that of the teacher rather than relying on absolute feature positions or isolated point-wise imitation.
Furthermore, DFM-KD is lightweight, architecture-agnostic, and can be seamlessly integrated into a wide range of Transformer-based trackers.
Our main contributions are:
\begin{itemize}
\item We introduce a novel distillation mechanism that transfers the teacher’s batch-wise topology by aligning inter-sample difference feature maps, capturing structural knowledge ignored by instance-wise approaches.
\item DFM-KD is flexible and can be selectively applied to arbitrary Transformer layers, enabling practitioners to target the most informative layers while minimizing overhead.
\item We demonstrate that DFM-KD integrates effortlessly with modern Transformer-based trackers (e.g., STARK, MixFormerV2) and significantly improves tracking precision and success rate with minimal computational cost.
\end{itemize}

\section{Related Work}

\subsection{Visual Object Tracking}
Visual Object Tracking (VOT) aims to estimate the position of a target object across video frames and is a fundamental perception task in autonomous driving and surveillance. Traditional tracking algorithms range from correlation filter methods to Siamese-based networks, achieving a good balance between accuracy and efficiency \cite{wu2015object,bertinetto2016fully,li2019siamrpn++,guo2020siamcar}. Recently, Transformer-based architectures have gained prominence due to their strong capability in modeling long-range dependencies and expressive feature representations \cite{yan2021learning} \cite{ye2022ostrack}. Methods such as STARK \cite{yan2021learning} and MixFormerV2 \cite{cui2023mixformerv2} utilize attention mechanisms to capture intricate spatiotemporal relations, setting new performance benchmarks on various tracking datasets.

Beyond these models, several recent Transformer-based trackers further advance interaction modeling by introducing stronger temporal and contextual fusion mechanisms. For instance, Liang \etal \cite{liang2025autoregressive} and Guo \etal \cite{guo2025dreamtrack} enrich cross-frame attention with autoregressive or future-aware temporal cues, while Li \etal \cite{li2025dynamic} and Videnovic \etal \cite{videnovic2025distractor} enhance hierarchical fusion through dynamic language--vision alignment or distractor-aware memory updates. In parallel, the expert-driven architecture in \cite{cai2025spmtrack} improves relation modeling within attention blocks by routing diverse spatial-temporal patterns to specialized experts.
Collectively, these works reflect a clear trend: state-of-the-art VOT performance is increasingly driven by stronger Transformer interaction layers and deeper hierarchical feature fusion. However, such Transformer-based designs are computationally expensive, and effectively transferring their rich interaction-layer knowledge into lightweight student models remains largely unaddressed—a natural opportunity for distillation-based approaches.

\subsection{Knowledge Distillation for Efficient Tracking}
Knowledge distillation (KD) is widely employed to compress high-capacity teacher models into lightweight student networks, aiming to retain accuracy while improving inference efficiency \cite{hinton2015distilling}. In the context of visual tracking, existing KD approaches generally align teacher and student feature representations on a per-instance basis \cite{11097763}. Early distillation methods \cite{zhu2017feature,dunnhofer2020tracking} improve efficiency by training lightweight trackers to mimic either the teacher’s intermediate feature representations or its final bounding-box predictions.
However, these approaches largely overlook inter-sample relational knowledge that captures the teacher’s batch-wise feature topology—an important source of robustness in dynamic tracking scenarios.

Our method addresses this gap by transferring inter-sample difference feature maps to better align the student's learned feature manifold structure with that of the teacher.
This distillation-based strategy offers a stable and parameter-efficient alternative to other efficiency paradigms. For instance, while dynamic methods exemplified by SGLATrack \cite{xue2025similarity} achieve efficiency by dynamically skipping layers, this approach requires retaining full model parameters and can introduce potential latency fluctuations when encountering complex frames.

\subsection{Relational and Sample-Relation Knowledge Distillation}

Conventional knowledge distillation primarily follows an instance-wise paradigm, training the student to mimic each teacher sample independently. In contrast, relational or sample-relation distillation focuses on how a teacher organizes samples within its embedding space. Representative methods include Relational Knowledge Distillation (RKD) \cite{park2019rkd}, which matches pairwise distances and angles; Correlation Congruence (CCKD) \cite{peng2019cckd}, which aligns batch-level correlation matrices; and Similarity-Preserving KD (SPKD) \cite{tung2019spkd}, which enforces pairwise similarity consistency.
Relational ideas have also appeared in other domains. Prototype–Sample Relation Distillation (PRD) \cite{asadi2023prd} preserves prototype–sample similarity structures in continual learning, and relation-based distillation has been applied to object detection \cite{relationKD_det} and source-free domain adaptation \cite{hier_rel_distill}. Virtual Relation Matching (VRM) \cite{zhang2025vrm} further generalizes relational knowledge transfer through a learned relation encoder.

Our DFM-KD follows the spirit of relational KD but differs in how relations are constructed. Instead of building full pairwise matrices, we compute lightweight inter-sample difference feature maps along the batch dimension. These differences capture how teacher representations vary across samples—encoding appearance changes, contextual similarity, and structural relationships—while keeping the computational and memory cost linear in batch size. This provides complementary guidance to feature regression and is particularly effective for transformer-based tracking, where robust separation among different targets and between targets and distractors is critical.

\section{Methodology}
\begin{figure*}[h]
  \centering
  \includegraphics[width=0.41\linewidth]{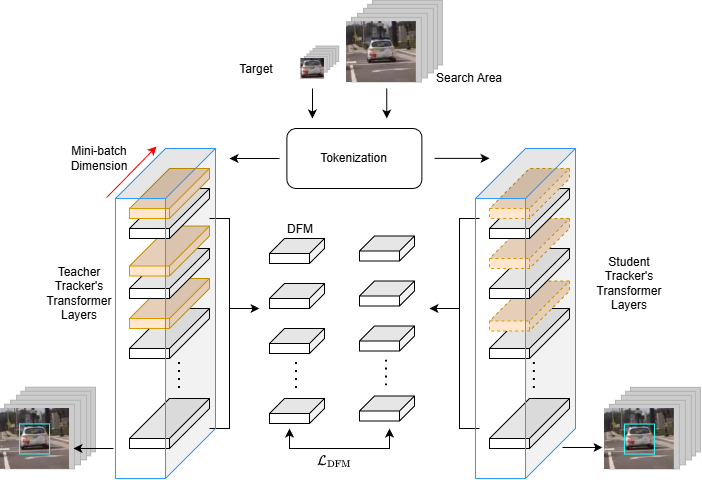}
  \includegraphics[width=0.41\linewidth]{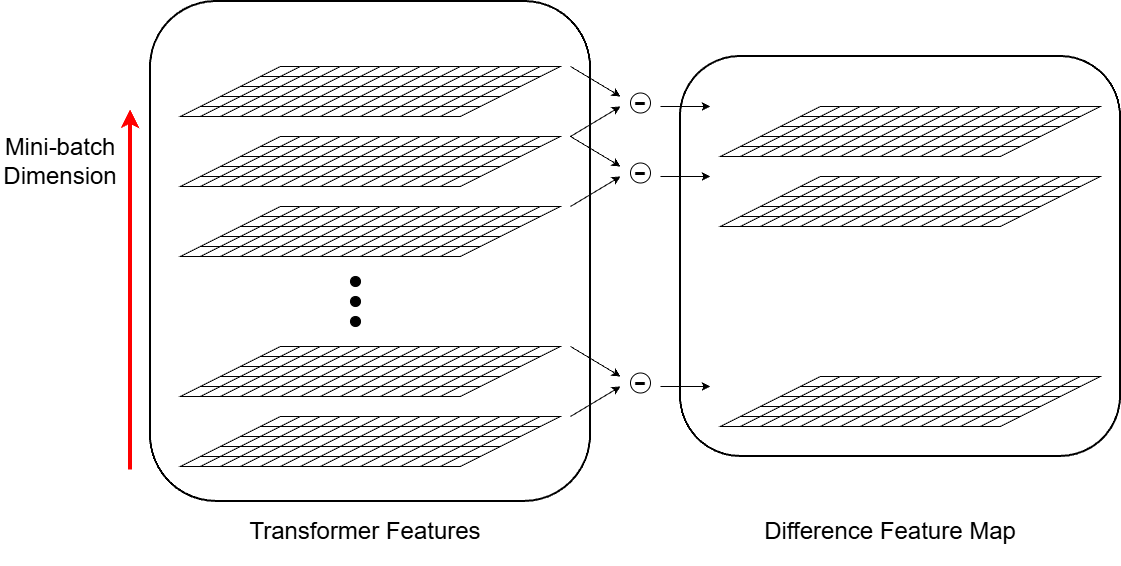} 
  \caption{Illustration of the Difference Feature Map based distillation. Note: Grey cubes are selected for knowledge transfer. Dashed yellow cubes that are eliminated do not participate in distillation. The right panel provides a detailed view of how difference feature maps are constructed along the mini-batch dimension.} 
  \label{fig:DFM} 
\end{figure*}



\subsection{Baseline Feature-Level Distillation}
\label{subsec:dfm}

In conventional feature-level knowledge distillation for visual recognition and detection \cite{yim2017gift}\cite{zagoruyko2016paying}, the student network is trained to directly regress the teacher's feature maps on a per-sample basis, e.g., by minimizing the mean squared error (MSE) between the teacher and student activations at selected layers. Formally, for a given transformer layer $\ell$, let
\begin{equation}
    \mathbf{F}^{T}_{\ell} \in \mathbb{R}^{B \times N \times C}, 
    \qquad
    \mathbf{F}^{S}_{\ell} \in \mathbb{R}^{B \times N \times C}
\end{equation}
denote the teacher and student feature tensors, where $B$ is the mini-batch size (each element corresponds to one training sample, e.g., a tracking pair(template, search area) or triplet(template, online template, search area)), $N$ is the number of tokens (template and search tokens combined), and $C$ is the channel dimension (transformer embedding size). A standard feature-distillation loss can be written as:
\begin{equation}
    \mathcal{L}^{(\ell)}_{\text{feat}} 
    = \frac{1}{B N C} 
      \sum_{b=1}^{B} 
      \sum_{n=1}^{N} 
      \sum_{c=1}^{C}
      \left(
        F^{S}_{\ell, b, n, c} - F^{T}_{\ell, b, n, c}
      \right)^{2},
    \label{eq:feat_kd}
\end{equation}
where $F^{T}_{\ell, b, n, c}$ and $F^{S}_{\ell, b, n, c}$ denote the individual scalar entries of $\mathbf{F}^{T}_{\ell}$ and $\mathbf{F}^{S}_{\ell}$, respectively. While this sample-wise alignment is effective, it treats each sample in the batch independently and does not preserve the teacher’s latent organization across samples, which can limit robustness under challenging tracking conditions.

\subsection{Difference Feature Map Distillation}
In visual object tracking, we observe that a well-trained teacher does more than generate discriminative features for individual samples. It also establishes a meaningful batch-wise topology. Easy, hard, and outlier examples naturally occupy distinct regions and relative distances within the teacher’s embedding space. Ignoring this structure forces the student to imitate features in isolation, which can lead to a myopic form of matching that accelerates convergence but discards valuable inter-sample relationships.
To overcome this limitation, we introduce Difference Feature Map Distillation (DFM-KD), a relational distillation strategy that transfers how the teacher’s representations change across samples within mini-batches, rather than only supervising the absolute features of each sample. By explicitly distilling these inter-sample differences, DFM-KD preserves the teacher’s latent topology and provides richer structural guidance than conventional point-wise feature matching. Fig.~\ref{fig:DFM} provides an overview of DFM-based distillation.

\subsubsection{Construction of difference feature maps}

As before, for a given transformer layer $\ell$, we denote the teacher and student feature tensors as:
\begin{align}
    \mathbf{F}^{T}_{\ell} &= 
    \{ F^{T}_{\ell, b, n, c} \}_{b=1,n=1,c=1}^{B, N, C}
    \in \mathbb{R}^{B \times N \times C}, \\
    \mathbf{F}^{S}_{\ell} &= 
    \{F^{S}_{\ell, b, n, c} \}_{b=1,n=1,c=1}^{B, N, C}
    \in \mathbb{R}^{B \times N \times C}.
\end{align}
Here, the batch index $b$ enumerates different training samples within the mini-batch, where each sample may correspond to a tracking pair or triplet composed of template and search regions. The token index $n$ enumerates all tokens associated with a given sample, including both template and search tokens, and the channel index 
$c$ indexes the embedding channels of the transformer.

We construct \emph{difference feature maps} by taking discrete differences along the batch dimension. Specifically, for the teacher features, we define the difference feature map $\mathbf{D}^{T}_{\ell} \in \mathbb{R}^{(B-1) \times N \times C}$ with entries:
\begin{equation}
\label{eq:teacher_diff}
D^{T}_{\ell,i,n,c}
=
F^{T}_{\ell,i+1,n,c}
-
F^{T}_{\ell,i,n,c},
\qquad i = 1,\dots,B-1.
\end{equation}
In other words, we subtract the feature of sample $i$ from that of sample $i+1$ for every token and channel, yielding a set of $(B-1)$ difference maps that capture how the teacher’s representations vary between neighboring samples in the mini-batch.
Analogously, the student difference tensor $\mathbf{D}^{S}_{\ell} \in \mathbb{R}^{(B-1) \times N \times C}$ is defined as:
\begin{equation}
\label{eq:student_diff}
D^{S}_{\ell,i,n,c}
=
F^{S}_{\ell,i+1,n,c}
-
F^{S}_{\ell,i,n,c},
\qquad i = 1,\dots,B-1.
\end{equation}

Equations~\eqref{eq:teacher_diff}–\eqref{eq:student_diff} can be interpreted as forming an ``ordered batch-wise derivative'' along the sample dimension, capturing how representations evolve across samples. They encode appearance variations, contextual similarities, and structural relationships.
Instead of only considering each feature $\mathbf{F}_{\ell, b,:,:}$ in isolation like conventional feature-level distillation does, we explicitly model how features vary from one sample to the next within a batch.
In practice, we follow the standard mini-batch ordering as provided by the data loader, which corresponds to randomly shuffled training samples. This randomization implies that the differences encode generic inter-sample relationships, rather than being tied to a specific temporal order in a sequence.

We also note that DFM-KD operates on the output token embeddings of the transformer layer, regardless of whether the underlying architecture has an explicit encoder-decoder structure (e.g., STARK\cite{yan2021learning}) or a pure transformer stack (e.g., MixFormerV2\cite{cui2023mixformerv2}). As long as intermediate transformer layer outputs are in the form of token embeddings, the difference feature maps can be computed as in~\eqref{eq:teacher_diff}--\eqref{eq:student_diff}.

In principle, one could construct pairwise differences for all sample pairs within the batch, i.e., $F_{\ell, b_2,:,:} - F_{\ell, b_1,:,:}$ for all $1 \leq b_1 < b_2 \leq B$, which would yield $O(B^2)$ difference maps per layer. However, this quadratic construction is computationally expensive and memory-intensive for typical batch sizes in transformer-based tracking. Instead, we compute differences only between adjacent samples $(i, i+1)$, yielding exactly $(B-1)$ maps and reducing the complexity to $O(B)$. Coupled with the random shuffling of mini-batch composition, this “adjacent difference’’ approximation offers an effective trade-off: it remains computationally lightweight while still capturing meaningful inter-sample topology.

\subsubsection{DFM loss}

Given the teacher and student difference feature maps, $\mathbf{D}^{T}_{\ell}$ and $\mathbf{D}^{S}_{\ell}$, we define the DFM loss at layer $\ell$ as the mean squared error between the two tensors:
\begin{equation}
    \mathcal{L}^{(\ell)}_{\text{DFM}}
    =
    \frac{1}{(B-1) N C}
    \sum_{i=1}^{B-1}
    \sum_{n=1}^{N}
    \sum_{c=1}^{C}
    \left(
        D^{S}_{\ell, i, n, c}
        -
        D^{T}_{\ell, i, n, c}
    \right)^{2}.
    \label{eq:dfm_layer_loss}
\end{equation}

We would like to point out that we do not apply normalization to $\mathbf{D}^{T}_{\ell}$ and $\mathbf{D}^{S}_{\ell}$ before computing the loss in~\eqref{eq:dfm_layer_loss}. Normalizing each difference vector to unit length would remove the magnitude information of inter-sample variations and mainly preserve direction. However, the core motivation of DFM-KD is to transfer not only the direction, but also the relative strength of the teacher's feature changes across different samples. 
We found that retaining this magnitude information is important for preserving the teacher's batch-wise topology. Therefore, all experiments are conducted with raw difference feature maps.

When distilling multiple transformer layers, we aggregate the per-layer DFM losses. Let $\mathcal{L}_d$ denote the index set of transformer layers selected for DFM (e.g., student encoder layers aligned with specific teacher layers), and let $\alpha_{\ell}$ be a non-negative weighting factor for each layer. The overall DFM loss is then defined as:
\begin{equation}
    \mathcal{L}_{\text{DFM}}
    =
    \sum_{\ell \in \mathcal{L}_d}
    \alpha_{\ell} \,
    \mathcal{L}^{(\ell)}_{\text{DFM}}\,.
    \label{eq:dfm_total}
\end{equation}


\subsubsection{Overall Training Objective}
The DFM loss is combined with the standard tracking loss and the conventional feature-level distillation loss to form the overall training objective:
%
\begin{equation}
    \mathcal{L}_{\text{total}}
    =
    \mathcal{L}_{\text{track}}
    +
    \lambda_{\text{feat}} \,
    \mathcal{L}_{\text{feat}}
    +
    \lambda_{\text{DFM}} \,
    \mathcal{L}_{\text{DFM}},
    \label{eq:total_loss}
\end{equation}
where $\mathcal{L}_{\text{track}}$ denotes the original tracking supervision, $\mathcal{L}_{\text{feat}}$ is the standard feature distillation loss as defined in~\eqref{eq:feat_kd}, and $\lambda_{\text{feat}}$ and $\lambda_{\text{DFM}}$ are scalar weights controlling the contributions of the two distillation terms.

\section{Experiments and Results}



\subsection{Experimental Setup}
\subsubsection{Models}
To demonstrate the efficacy and generality of our method, we evaluate DFM-KD on two prominent and structurally distinct trackers: STARK \cite{yan2021learning} and MixFormerV2 \cite{cui2023mixformerv2}. STARK employs a CNN backbone (ResNet-50) to extract template and search features, which are then processed jointly through an encoder–decoder Transformer. In contrast, MixFormerV2 uses a pure transformer design with ViT-style patch embeddings and a stack of self-/cross-attention blocks operating directly on tokens. Despite their architectural differences in backbone design and token formation, both trackers share a common structural pattern — a sequence of Transformer layers that progressively refine token embeddings. This shared structural pattern allows our distillation method to be applied with minimal adaptation, highlighting the versatility of DFM-KD across heterogeneous tracking architectures.
For each tracker, we construct a lightweight student model by reducing the number of transformer layers (e.g., STARK: 12$\rightarrow$6 layers; MixFormerV2: 12$\rightarrow$8 layers). Teacher and student layers are aligned in a straightforward manner (e.g., by selecting evenly spaced teacher layers), and distillation is performed directly on the corresponding transformer outputs. We revisit this fixed layer pairing as a future direction in Section~\ref{sec:future}.

\subsubsection{Data}
We conduct our experiments on the LaSOT2019 benchmark \cite{fan2019lasot}, a large-scale long-term tracking dataset containing 1,400 sequences. Although LaSOT2019 includes a total of 70 categories, our focus is on the agents that function as typical road users in autonomous driving scenarios—those that interact with or move directly within the roadway. To this end, we construct a dedicated subset, denoted LaSOT-RoadUsers, consisting of six categories: bicycle, motorcycle, car, bus, truck and person.
All evaluations follow the official LaSOT protocol, using AUC, Precision (P), and Normalized Precision (NP).

\subsubsection{Training Details}
Our experiments were conducted on two NVIDIA A100 GPUs. For the STARK architecture, we adopted the teacher model from the MMTracking repository \cite{mmtrack2020}, which consists of a 12-layer Transformer with a ResNet-50 backbone. In this setting, the hyperparameters $\lambda_{\text{feat}}$ and $\lambda_{\text{DFM}}$ were set to 0.2 and 0.5, respectively. For the MixFormerV2 architecture, we employed the official 12-layer, 288-base model as the teacher and incorporated DFM-KD to distill an 8-layer student model, following the MixFormerV2 distillation protocol. In this case, $\lambda_{\text{feat}}$ and $\lambda_{\text{DFM}}$ were set to 0.2 and 0.3, respectively.

\subsection{Quantitative Results}

\begin{table}[h]
\centering
\caption{Comparison of methods on MixFormerV2.}
\begin{tabular}{lccccc}
\toprule
\textbf{Method} & \textbf{Params(M)} & \textbf{FLOPs(G)} & \textbf{AUC} & \textbf{P} & \textbf{NP}  \\ 
\midrule
Teacher &86.60&42.10 &75.33&81.10&86.66\\
\midrule
MixFormerV2  & \multirow{3}{*}{58.25} & \multirow{3}{*}{28.16} & 73.21  & 77.90  & 84.27  \\  
TDSC-KD    &  & & 73.78  & 78.39  & 84.74  \\  
DFM-KD    &  & &74.03	& 78.85 & 85.25 \\  
\bottomrule
\end{tabular}
\label{tab:result} 
\end{table}

\begin{table}[h]
\centering
\caption{Comparison of methods on STARK. FLOPs are computed for the transformer component only.}
\begin{tabular}{l@{\hspace{3pt}}ccccc}
\toprule
\textbf{Method} & \textbf{Params(M)} & \textbf{FLOPs(G)}& \textbf{AUC}  & \textbf{P} & \textbf{NP}  \\ 
\midrule
Teacher      & 27.87&5.80   &68.63 &72.91 &79.95\\
\midrule
Student-baseline   & \multirow{3}{*}{19.19}& \multirow{3}{*}{3.29}      &64.20	&62.77	&72.07  \\ 
Conventional KD       & &  &65.94	&67.98	&74.59 \\ 
DFM-KD    & & &68.76	&72.17	&78.65  \\ 
\bottomrule
\end{tabular}
\label{tab:result1}
\end{table}



Table~\ref{tab:result} summarizes the results on MixFormerV2. The vanilla 8-layer student exhibits a clear performance drop relative to the 12-layer teacher, whereas TDSC-KD partially recovers this loss. Our DFM-KD achieves the strongest student performance, boosting NP by +0.98 over the vanilla student and by +0.51 over TDSC-KD.
On STARK (Table~\ref{tab:result1}), the student baseline lags behind the teacher by a large margin. DFM-KD yields substantially stronger improvements than both the student baseline and conventional KD, markedly narrowing the teacher–student gap and even surpassing the teacher in AUC.
Notably, DFM-KD achieves these improvements without increasing parameters or FLOPs over the student models, preserving computational efficiency at inference time.



Table~\ref{tab:result2} ablates the effects of DFM-only and vanilla feature KD (two components in Eq. \ref{eq:total_loss}). As we can see, both components improve the student baseline, with DFM-only providing a noticeably larger boost. When the two KD components are combined, the model achieves the strongest overall performance across all three metrics.

\begin{table}[h]
\centering
\caption{Ablation of DFM-KD components on STARK.}
\begin{tabular}{lcccc}
\toprule
\textbf{Method} & \textbf{AUC}  & \textbf{Precision} & \textbf{Norm Precision} \\ 
\midrule
Baseline         &64.20	&62.77	&72.07  \\ 
Feat-only         &65.94	&67.98	&74.59 \\ 
DFM-only          &67.81	&70.70	&77.75  \\ 
DFM-KD (combined)    &68.76	&72.17	&78.65  \\ 
\bottomrule
\end{tabular}
\label{tab:result2}
\end{table}


\subsection{Qualitative Results}
\begin{figure*}[!h]
  \centering
  \includegraphics[width=0.2\linewidth]{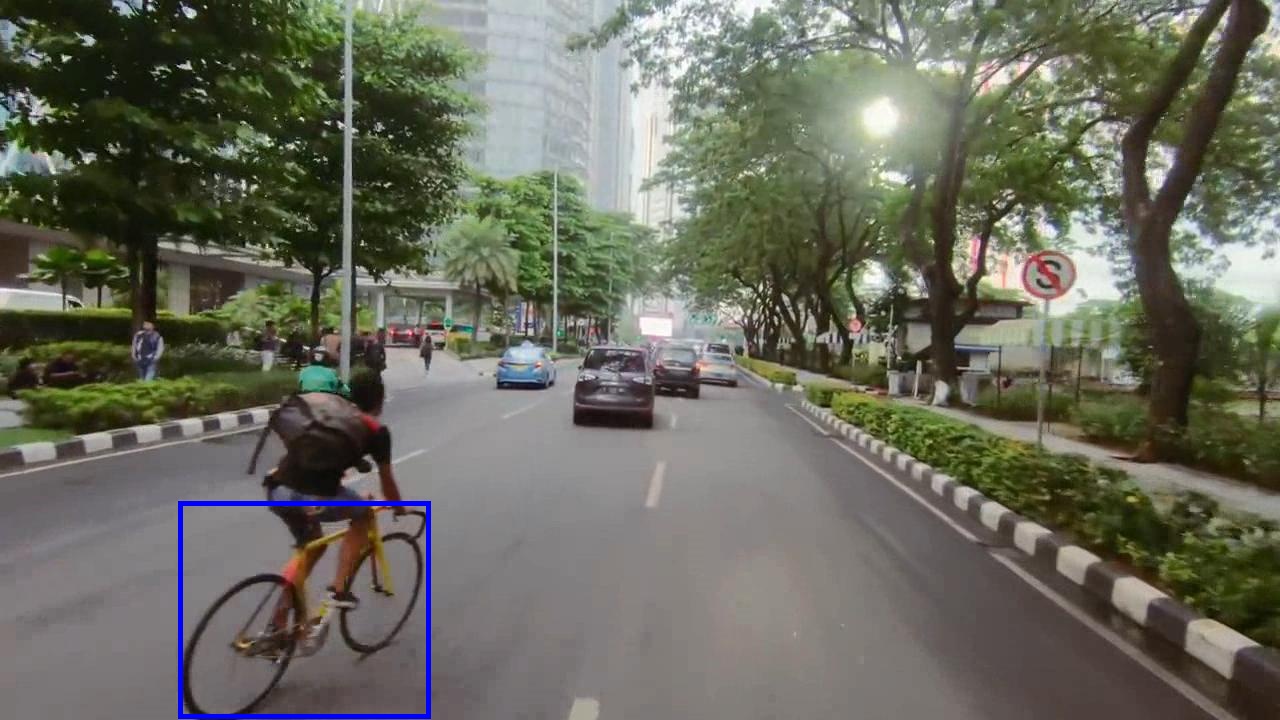}
  \raisebox{5ex}{\scalebox{1.5}{\textbf{\ldots}}}
  \includegraphics[width=0.2\linewidth]{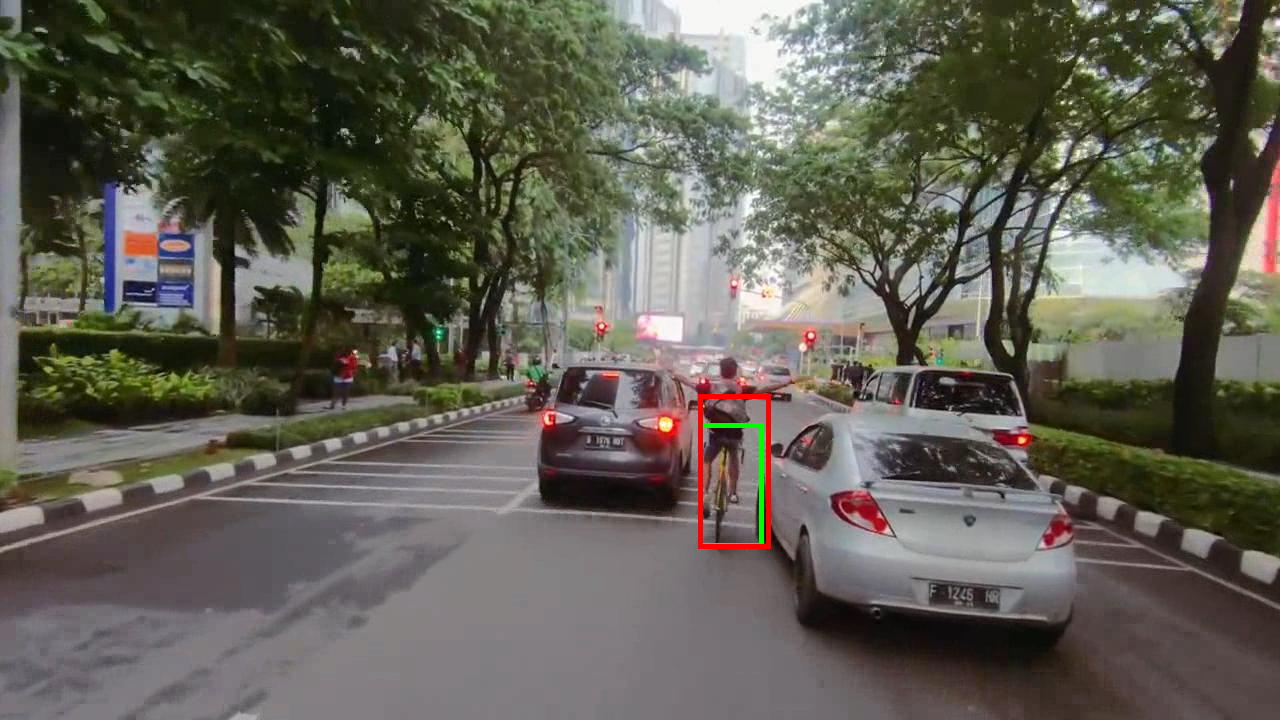}
  \includegraphics[width=0.2\linewidth]{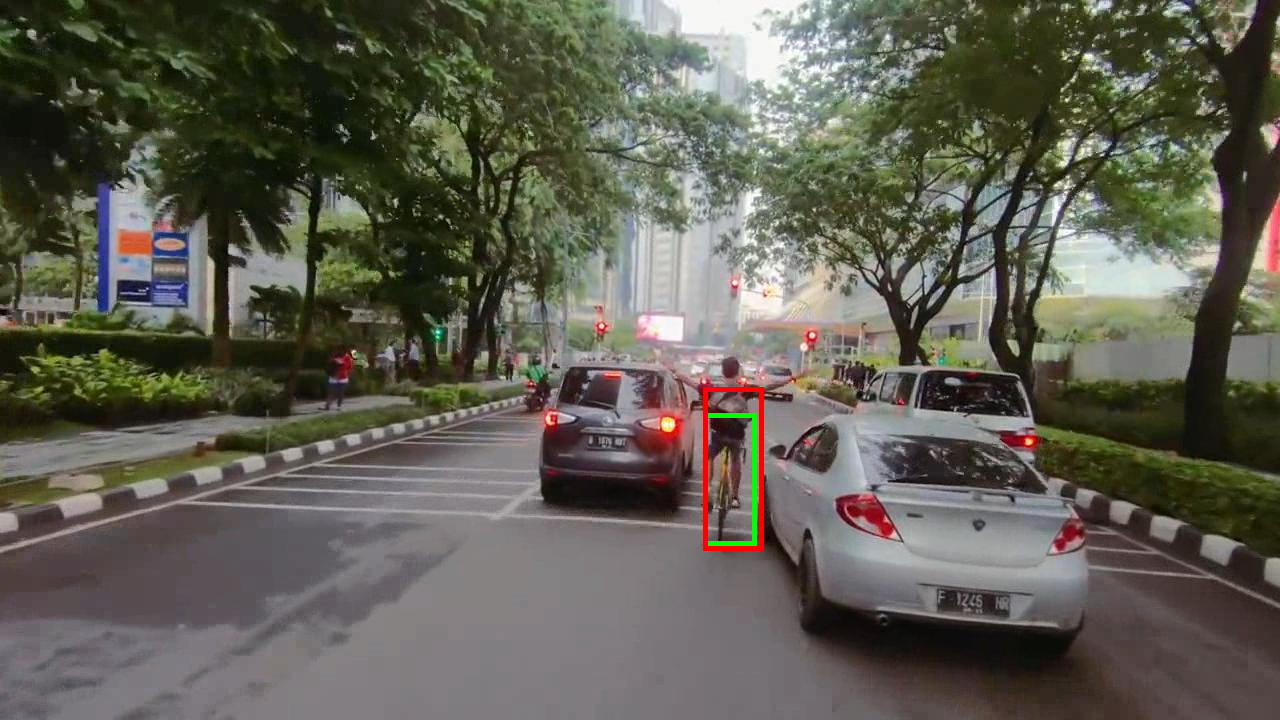}
  \includegraphics[width=0.2\linewidth]{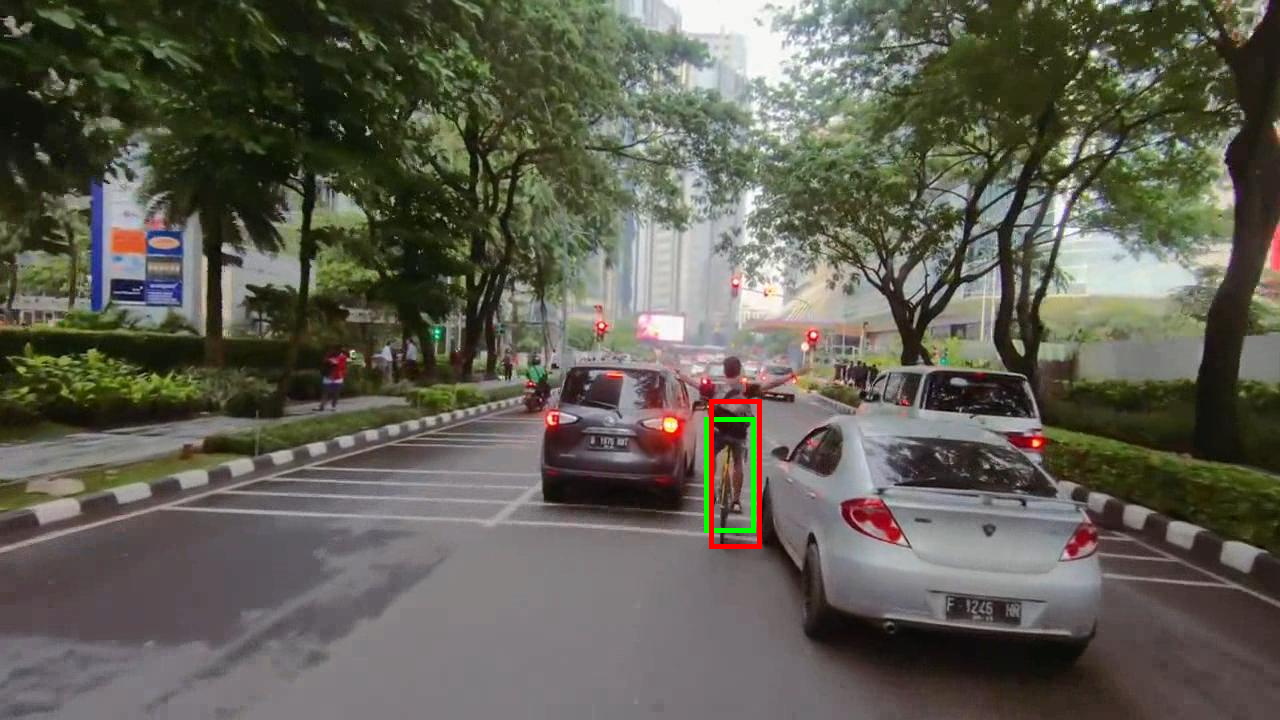}
  \includegraphics[width=0.2\linewidth]{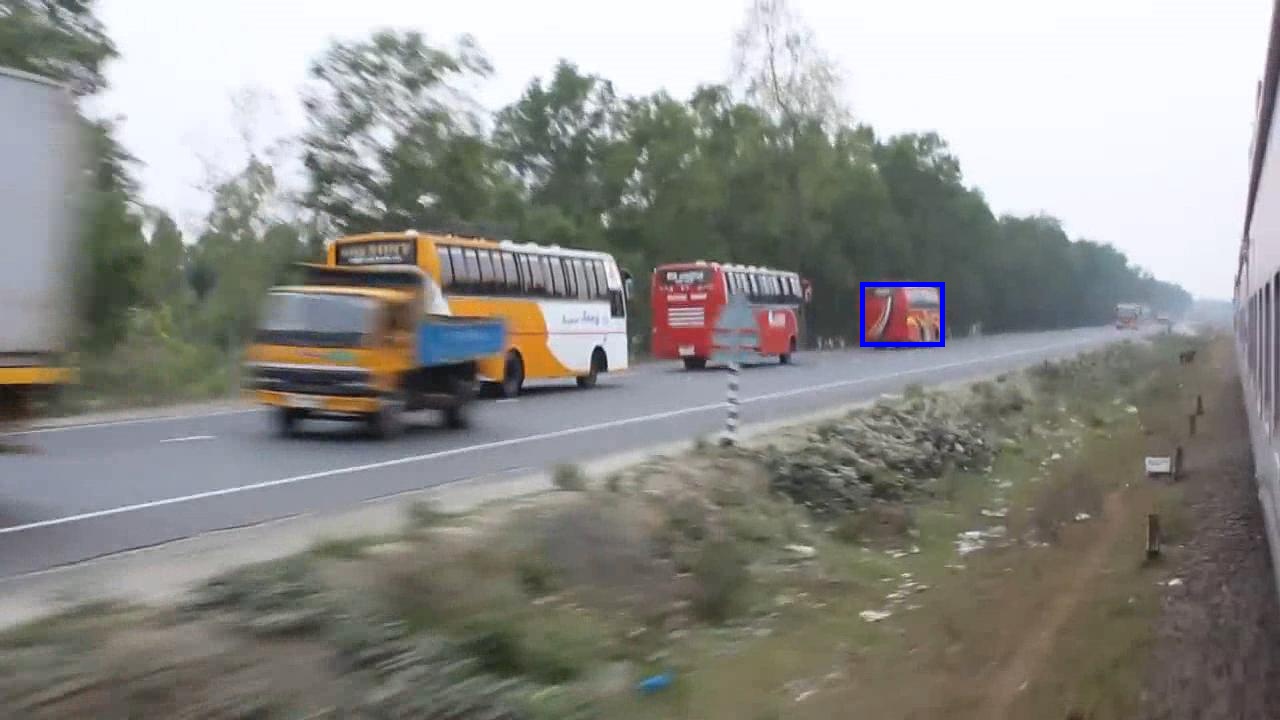}
  \raisebox{6ex}{\scalebox{1.5}{\textbf{\ldots}}}
  \includegraphics[width=0.2\linewidth]{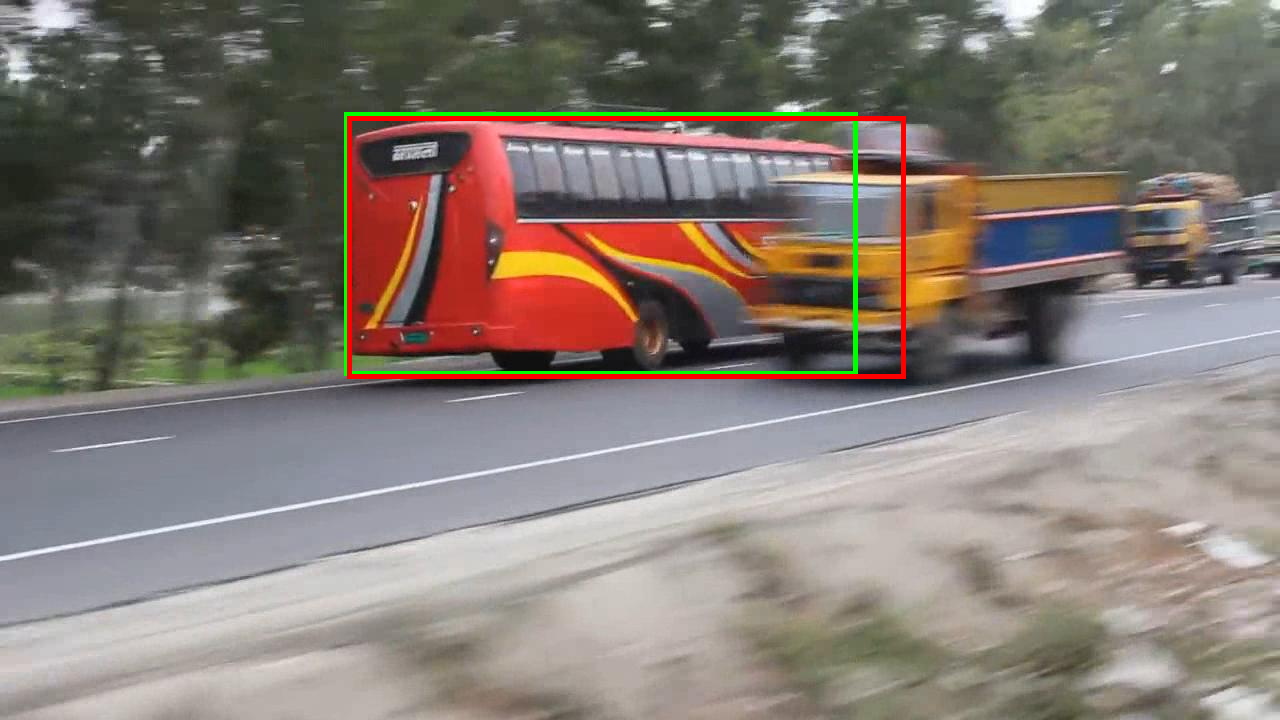}
  \includegraphics[width=0.2\linewidth]{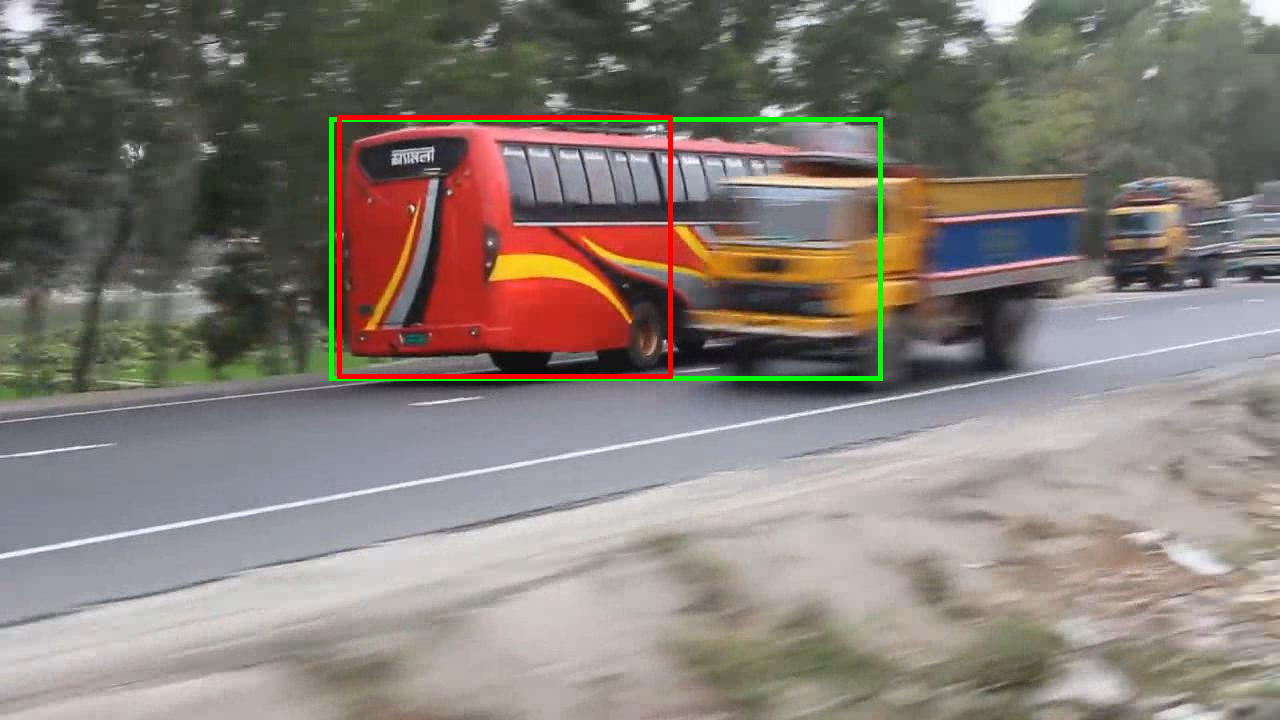}
  \includegraphics[width=0.2\linewidth]{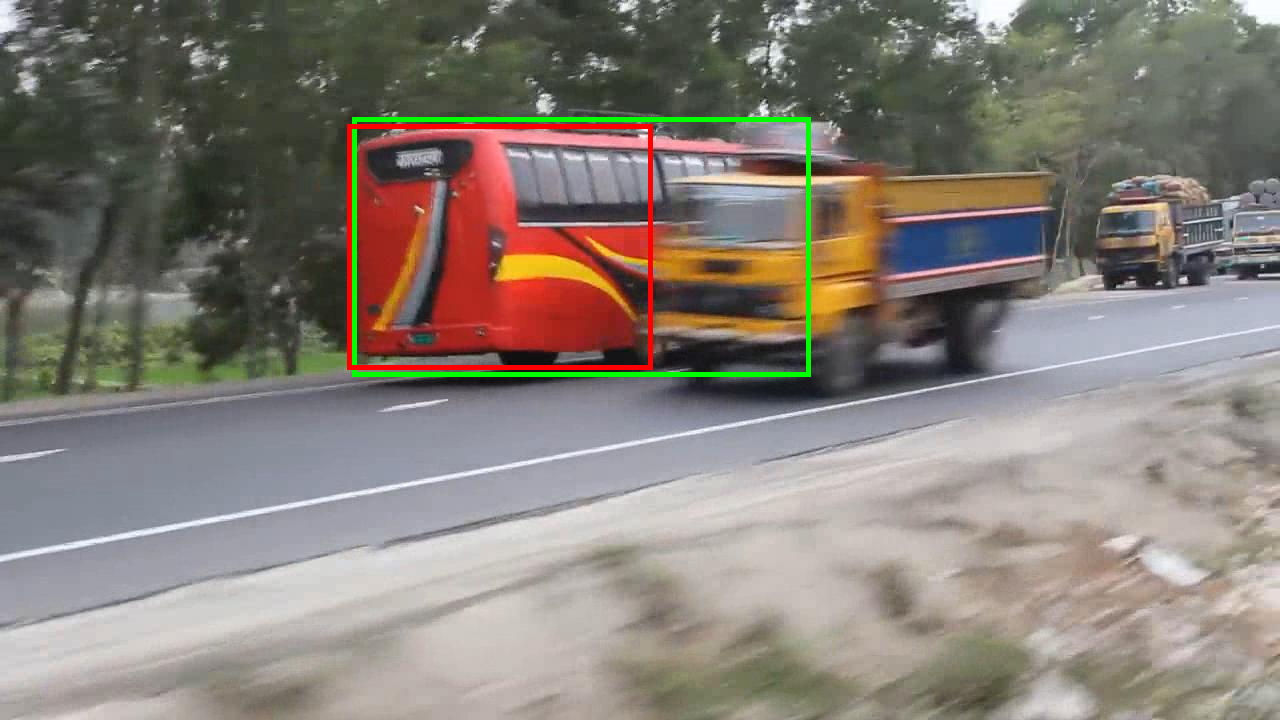}
  \includegraphics[width=0.2\linewidth]{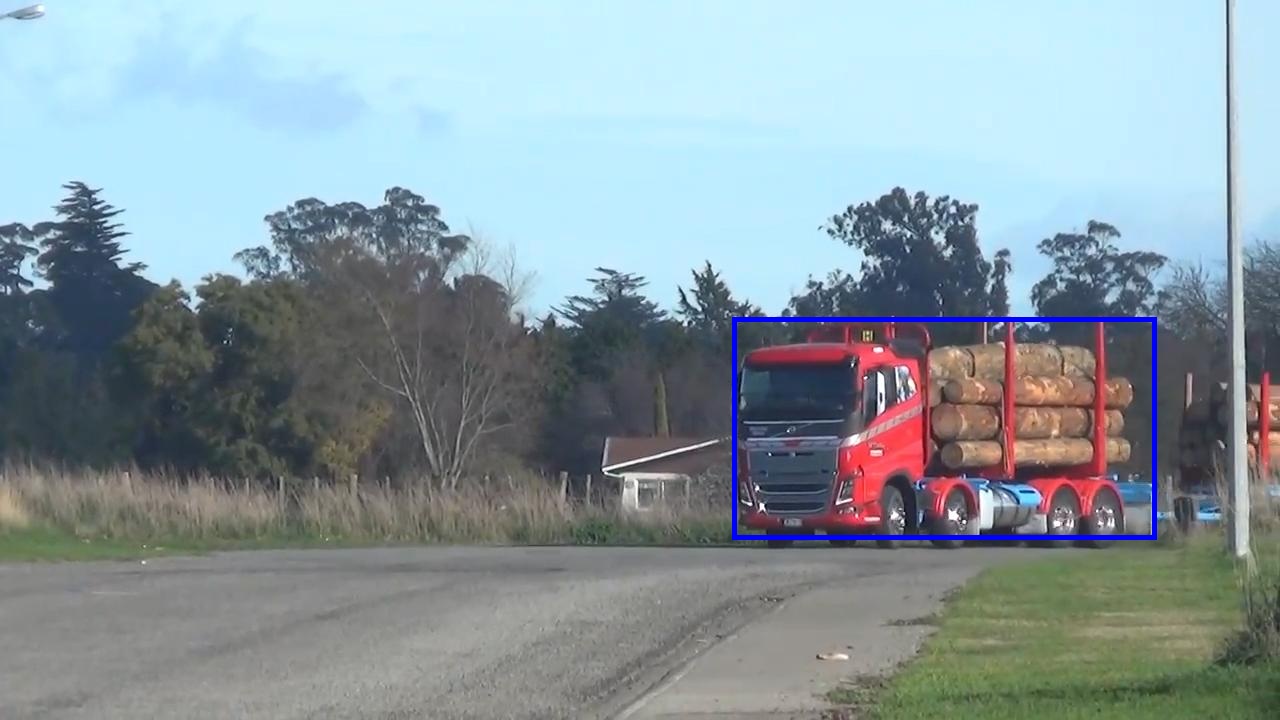} 
  \raisebox{7ex}{\scalebox{1.5}{\textbf{\ldots}}}
  \includegraphics[width=0.2\linewidth]{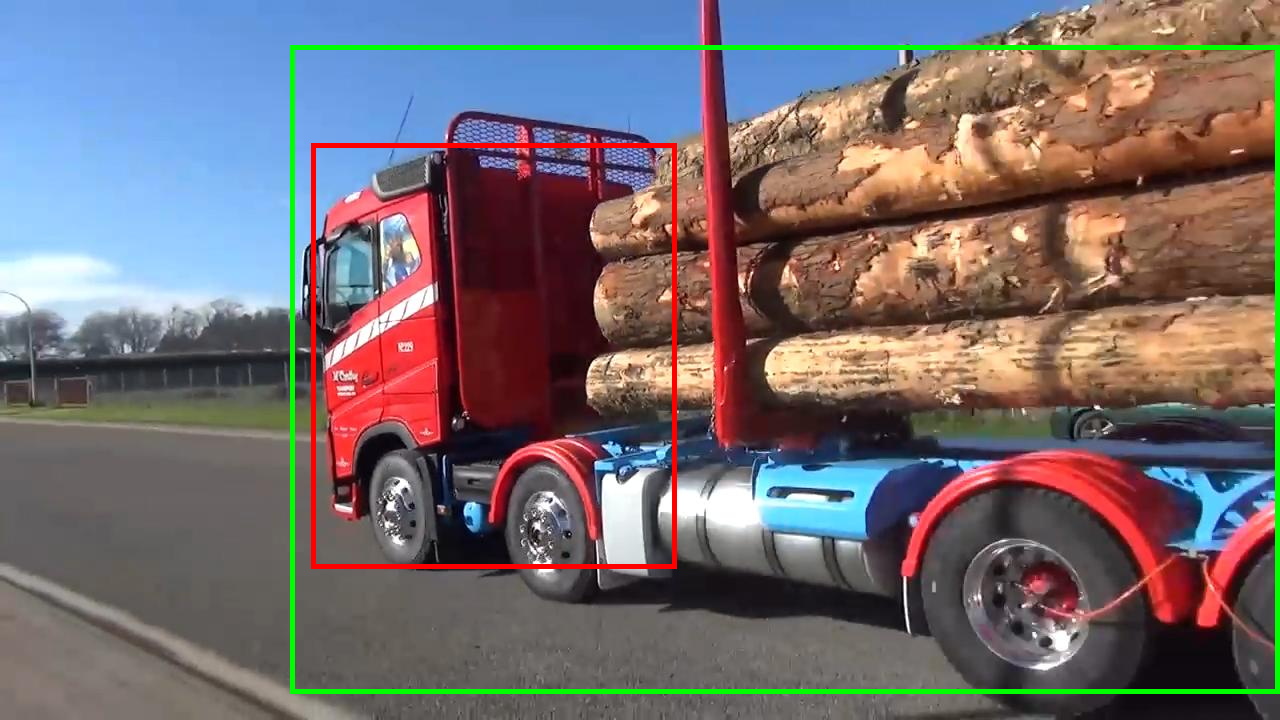}
  \includegraphics[width=0.2\linewidth]{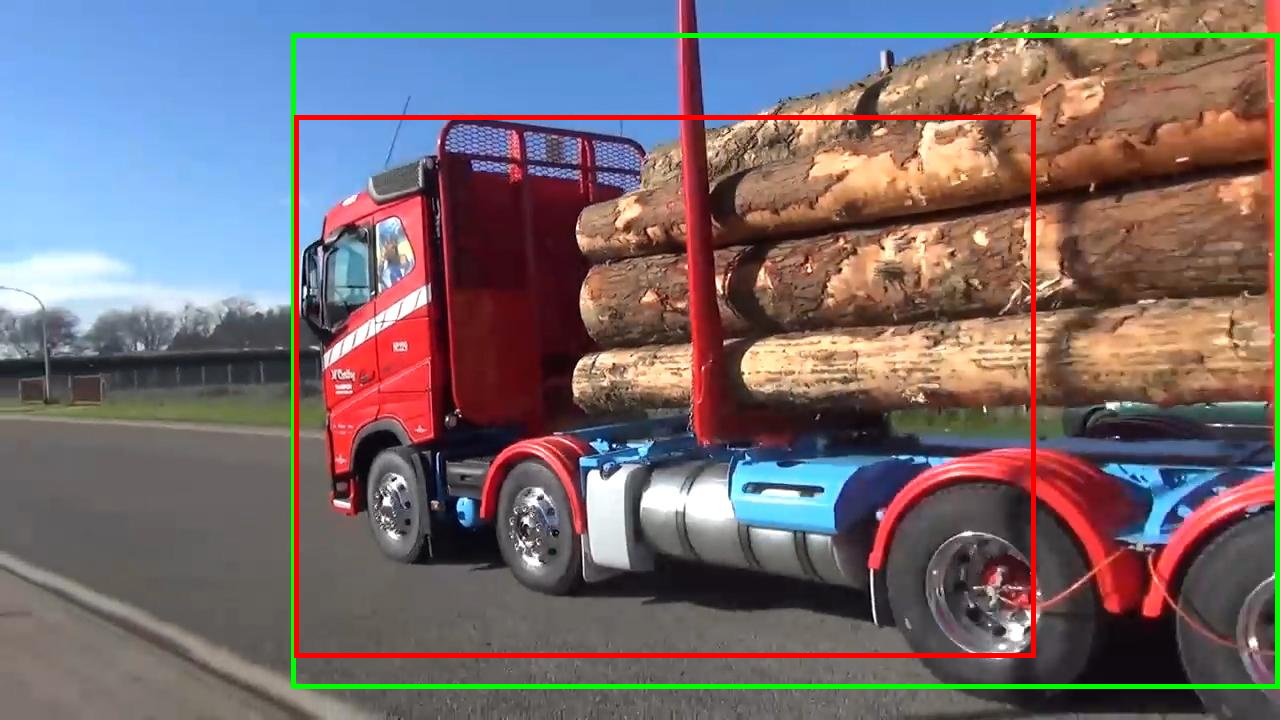}
  \includegraphics[width=0.2\linewidth]{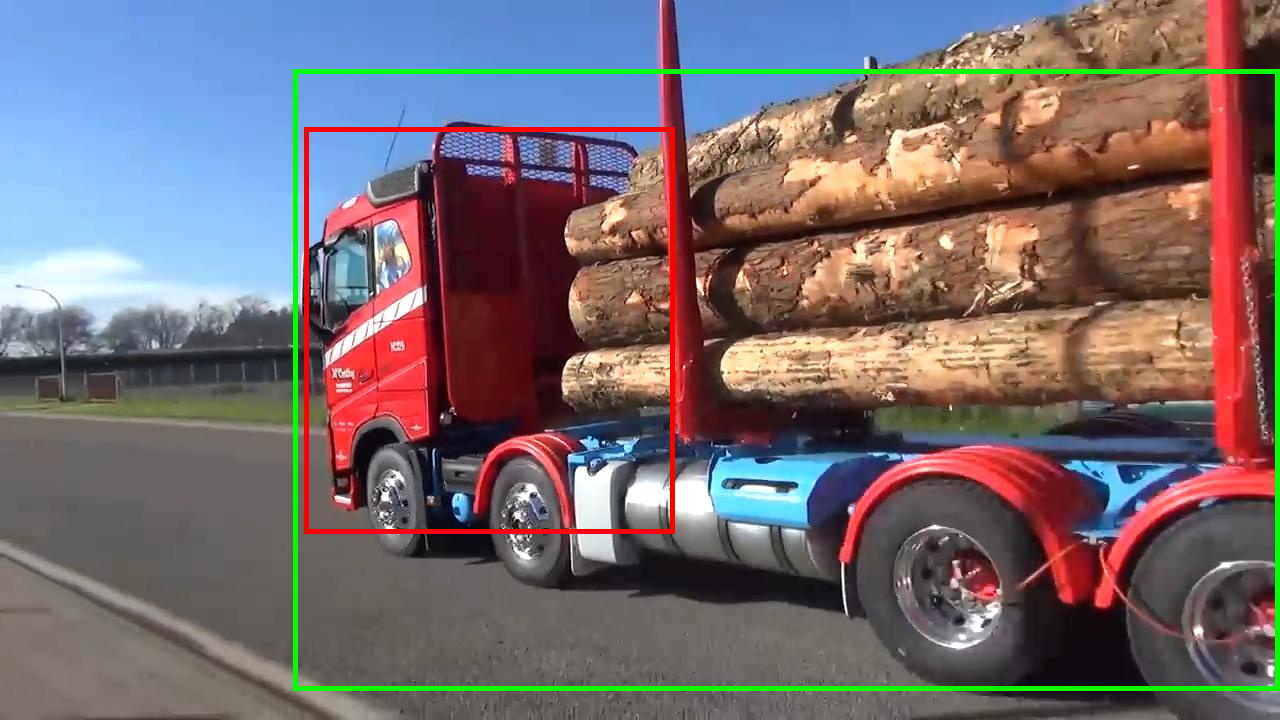}
  \caption{Qualitative tracking results. The dark blue bounding box marks the target in the initial frame. Across the following three frames, the green bounding box shows the output of our DFM-KD model, whereas the red bounding box corresponds to the student baseline (STARK architecture).} 
  \label{fig:plotresult} 
\end{figure*}

Figure \ref{fig:plotresult} demonstrates some example qualitative results. Each row illustrates a progression of frames from a video sequence, starting with the initial frame on the left and followed by three subsequent frames where the two trackers exhibit different behaviors. The target object is marked with a blue bounding box in the first frame, while the green and red boxes denote the predictions of our DFM-KD tracker and the student baseline (STARK architecture), respectively. Across these examples, our DFM-KD model consistently delivers more stable and accurate tracking under challenging conditions.

For instance, in the first row, our DFM-KD tracker maintains a tighter and more stable localization of the bicycle compared with the baseline. In the second row, the baseline model’s bounding box progressively shrinks as occlusion increases, ultimately failing to precisely keep track of the target. In contrast, DFM-KD preserves accurate localization throughout the sequence, benefiting from the transferred inter-sample relational structure. The third row presents another challenging scenario involving significant viewpoint changes: while the baseline shows clear bounding-box instability, our DFM-KD tracker produces consistently tighter and more accurate predictions. These examples suggest that distilling relational differences helps the student model preserve the teacher’s feature-space topology, leading to improved robustness under distractors, occlusions, and significant geometric variations.

\section{Conclusion}
In this work, we introduced Difference Feature Map Knowledge Distillation (DFM-KD), a relational distillation paradigm that transfers the teacher’s inter-sample topology by aligning difference feature maps across training samples within a batch, rather than relying solely on conventional point-wise feature regression. By explicitly modeling how representations change between neighboring samples, DFM-KD captures appearance variations, contextual similarities, and structural relationships that instance-wise distillation overlooks, while remaining computationally lightweight and architecture-agnostic. We demonstrated that DFM-KD can be seamlessly integrated into representative Transformer-based trackers, including STARK and MixFormerV2. Experiments on LaSOT-RoadUsers show that DFM-KD consistently improves tracking performance and outperforms state-of-the-art distillation approaches such as TDSC-KD, with ablation studies confirming its complementary benefits to standard feature-level distillation. These results emphasize the role of batch-wise manifold preservation in efficient visual tracking.

\section{Future Directions}
\label{sec:future}
A first direction is to extend relational distillation from the sample axis to the temporal axis. DFM-KD forms difference maps between neighboring samples within a batch, which captures how the teacher's representation varies across object instances but treats the frames of a video as unrelated draws. Tracking is intrinsically sequential, and the way a target's representation evolves across nearby frames encodes motion, deformation, and gradual appearance change that batch-wise differences cannot see. Temporal difference maps, computed between the teacher's features at temporally ordered frames of the same sequence (for instance between the initial template, the online template, and the current search region, or across a short window of search frames), would transfer how the teacher's representation responds to target dynamics. Aligning the student to these temporal differences, alongside the inter-sample ones, would preserve the teacher's feature-space trajectory over time rather than only its batch-wise topology. The template-to-search differences come at no extra data cost, since the trackers we distill already train on frames drawn from the same video, and the windowed variant would require only a modified frame sampler.

A second avenue concerns layer correspondence. In our current pipeline, which teacher layer each student layer imitates is fixed before training by an even-spacing rule and never revisited. Yet the teacher layer a given student layer would benefit from most is not obvious when the two networks differ in depth, and it may change as the student trains. Early on, a student layer may gain most from matching a nearby shallow teacher layer, while later it may be ready to absorb more abstract representations from deeper ones. A natural extension is therefore to make the correspondence itself learnable and student-aware: fusing teacher and student hidden states (e.g., via cross-attention) into a signal that reflects both the teacher's knowledge and the student's current state, and learning masks over the teacher's layer axis, and over the tokens within each layer, that re-weight every teacher--student pairing throughout distillation, with a diversity regularizer preventing collapse onto a single layer. This would replace the hand-chosen one-to-one alignment with an adaptive matching in which each student layer learns from a weighted mixture of teacher layers, with both the mixture and the emphasis within each layer evolving as the student does. The idea applies to any distillation setting where a deep teacher supervises a shallower student.


\bibliographystyle{plain} 
\bibliography{sample}

\end{document}